\documentclass[11pt]{article}

\usepackage[final]{acl}

\usepackage{times}
\usepackage{latexsym}
\usepackage{amsmath}
\usepackage{algorithm}
\usepackage{amssymb}
\usepackage{multicol, multirow}
\usepackage{booktabs}
\usepackage{array}
\usepackage{makecell}
\usepackage{url}
\usepackage{booktabs}
\usepackage{amsfonts}
\usepackage{nicefrac}
\usepackage{microtype}
\usepackage{xcolor}
\usepackage{graphicx}
\usepackage{amsmath}
\usepackage{amssymb}
\usepackage{makecell}
\usepackage{xspace}
\usepackage{algorithm,algpseudocode}
\usepackage{footmisc}
\usepackage{multirow}
\usepackage{array}
\usepackage{caption}
\usepackage{enumitem}

\usepackage{tabularx}
\usepackage{adjustbox}
\usepackage{enumitem} 
\usepackage{pifont} 
\usepackage{booktabs}
\newcommand{\cmark}{\ding{51}}%
\newcommand{\xmark}{\ding{55}}
\usepackage[T1]{fontenc}

\usepackage[utf8]{inputenc}

\usepackage{microtype}

\usepackage{inconsolata}

\usepackage{graphicx}

\title{Self-Evolving Dual-Role Deliberation for Medication Recommendation from Clinical Records}

\author{
\bf Jiyue Jiang$^{\heartsuit}$$^{\spadesuit}$$^{\ast}$,
Yu Lei$^{\heartsuit}$\thanks{These authors are co-first authors.},
Yitao Xu$^{\heartsuit}$,
Zikang Wang$^{\clubsuit}$,
Zhenyu Wang$^{\diamondsuit}$,\\
\bf Licheng Zong$^{\heartsuit}$,
Han Huang$^{\heartsuit}$,
Ziyi Li$^{\heartsuit}$,
Zhiyu Huo$^{\heartsuit}$, 
Yuhan Chen$^{\heartsuit}$, 
Jiayang Chen$^{\heartsuit}$,\\
\bf Xiaodan Fan$^{\heartsuit}$, 
Yu Li$^{\heartsuit}$\\
$^{\heartsuit}$ The Chinese University of Hong Kong,
$^{\spadesuit}$ Carnegie Mellon University,\\
$^{\clubsuit}$ The Hong Kong Polytechnic University, $^{\diamondsuit}$ Peking University Third Hospital\\
{\tt
\{jiangjy, yitaoxu, lczong, 1155269491, zhiyuhuo, yuhanchen, jiayangchen\}
} \\
{\tt
@link.cuhk.edu.hk,
leiyuu123@gmail.com, zikang.wang@connect.polyu.hk,
}\\
{\tt
1810301343@bjmu.edu.cn,
\{hanhuang, xfan\}@cuhk.edu.hk,
liyu@cse.cuhk.edu.hk
}
}

\begin{document}
\maketitle

\begin{abstract}
Recommending medications from electronic health records (EHRs) is a safety-critical task that demands therapeutic coverage, explicit safety constraints, traceable justification, and continual adaptivity. Supervised neural models freeze at deployment and cannot reason about new phenotypes, while existing LLM pipelines are single-pass and lack a cross-case improvement loop. We propose \textbf{SEMPA}, a training-free agentic framework that couples a \textsc{Proposer} with a \textsc{Critic} in multi-round adversarial deliberation, grounds candidates with molecular-aware retrieval, contrastively evolves both agents' natural-language policies under a semantic-anchor constraint, and enforces rule-based DDI and contraindication checks. On 100 MIMIC-IV encounters, SEMPA reaches $0.3116$ Jaccard, $0.7295$ F1, and a $0.0172$ DDI rate, improving F1 by $57.7\%$ and cutting DDI by $77.5\%$ over the strongest supervised baseline. Two LLM judges rank SEMPA highest, and six-backbone studies confirm that gains derive from the framework rather than a specific model. 
% Our code is available at \url{https://github.com/jiangjyjy/SEMPA}.
\end{abstract}

\section{Introduction}
\label{sec:intro}

Recommending medications from a patient's EHR is among the most consequential decision tasks in care \cite{wang2026clinical,johnson2023mimic}. A well-chosen regimen can resolve life-threatening pathophysiology in the intensive care unit; an ill-chosen one can precipitate adverse drug events, prolong hospitalization, or cause harm \cite{wishart2018drugbank,wang2026cancer}. Modern EHRs combine ICD-coded diagnoses, procedures, laboratory values, and free-text narratives, creating an opportunity to build assistive prescribing systems that augment clinician decision-making at the bedside \cite{jiang2025aibio}.

\begin{figure*}[t]
    \centering
    \includegraphics[width=\textwidth]{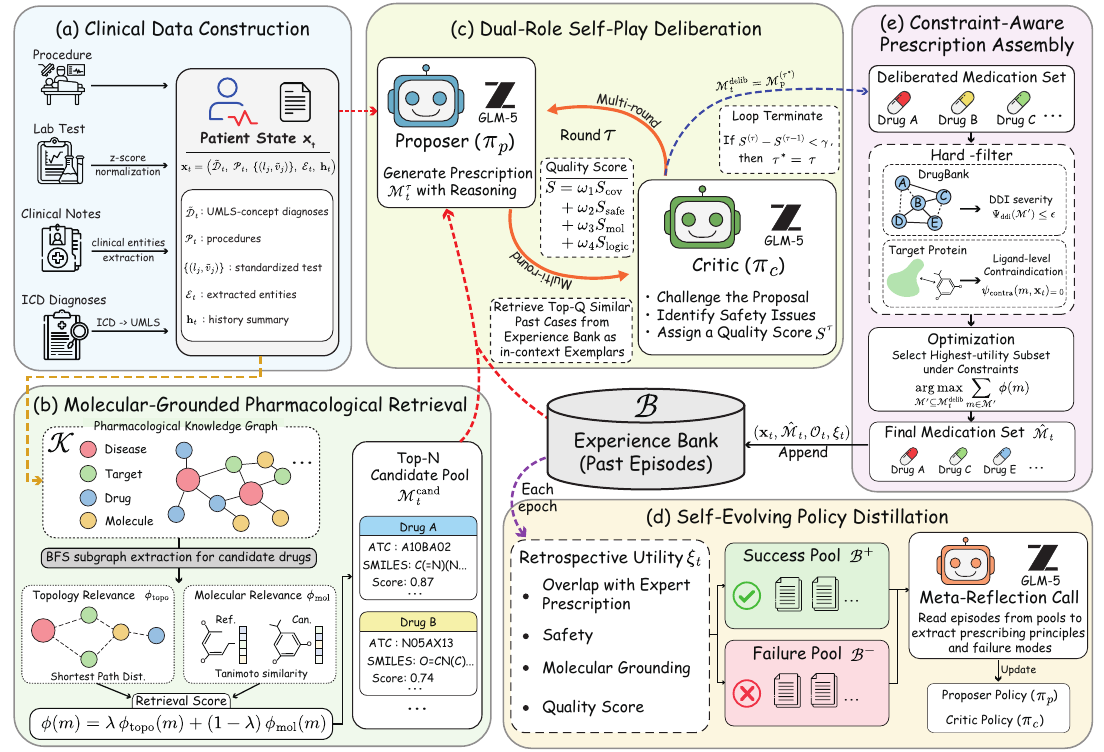}
    \caption{\textbf{Overview of SEMPA.} (a) \textbf{Clinical State Construction} normalizes EHRs into a patient state $\mathbf{x}_t$ (\S\ref{sec:m1}). (b) \textbf{Molecular-Grounded Retrieval} ranks candidates by a hybrid topology-molecule score (\S\ref{sec:m2}). (c) \textbf{Dual-Role Self-Play} alternates between a \textsc{Proposer} and a \textsc{Critic} over multiple rounds (\S\ref{sec:m3}). (d) \textbf{Self-Evolving Policy Distillation} contrastively rewrites agent policies (\S\ref{sec:m4}). (e) \textbf{Constraint-Aware Assembly} applies DDI and contraindication checks to yield $\hat{\mathcal{M}}_t$ (\S\ref{sec:m5}), which is appended to the experience bank $\mathcal{B}$ to close the loop.}
    \label{fig:overview}
\end{figure*}

The task is, however, difficult. Critical-care patients present with multiple co-occurring conditions, requiring multi-drug regimens that cover every active diagnosis while avoiding drug-drug interactions (DDIs), absolute contraindications, and organ-function violations such as nephrotoxic prescribing in patients with elevated creatinine \cite{yang2021safedrug,lu2020symptom}. The input is heterogeneous, the combinatorial drug space is large, and prescribing practice evolves as new evidence emerges, so any static decision system risks becoming stale once deployed. A clinically useful assistive system should reconcile four properties that are in tension: therapeutic \emph{coverage}, \emph{safety}, mechanistic \emph{justification}, and continual \emph{adaptivity}.

Two principal lines of work have addressed parts of this problem. Supervised models such as LEAP \cite{zhang2017leap}, RETAIN \cite{choi2016retain}, GAMENet \cite{shang2019gamenet}, SafeDrug \cite{yang2021safedrug}, MoleRec \cite{yang2023molerec}, and CausalMed \cite{li2024causalmed} learn drug-prediction functions via gradient-based optimization. While effective at capturing population-level patterns, they (i) freeze at deployment and cannot adjust to new phenotypes, (ii) treat prescribing as multi-label classification with no traceable rationale, (iii) lack molecular-level grounding for why a candidate fits the underlying pathophysiology, and (iv) often trade safety for coverage. More recently, LLMs have been applied via zero-shot prompting, chain-of-thought \cite{wei2022chain}, self-consistency \cite{wang2022selfconsistency}, retrieval-augmented generation \cite{lewis2020rag,xu2024ramehr,cao2025ehrrag}, and single-agent loops such as ReAct \cite{yao2023react} and Reflexion \cite{shinn2023reflexion}. These pipelines exploit the biomedical knowledge encoded in pretrained LLMs \cite{singhal2023large,wang2025biolm,jiang2025bench,jiang2025seq}, and related foundation-model studies have begun to extend such representations to biological sequences, RNA-protein interactions, and functional RNA design \cite{jiang2025rbp,jiang2026rmsa}, but existing clinical pipelines remain single-pass and single-perspective: a single model both proposes and critiques, with no explicit second opinion, no mechanism to learn from past prescribing successes and failures, and no explicit DDI constraint \cite{ke2025multiagent,huang2024mar}.

We argue that medication recommendation should be cast as an \emph{agentic, self-evolving deliberation process}, and we introduce \textbf{SEMPA} (Figure~\ref{fig:overview}). SEMPA's core, two LLM instances with role policies, a \textsc{Proposer} that drafts a prescription and a \textsc{Critic} that challenges it, engage in multi-round adversarial deliberation under a rubric. Around this core, three modules are added: a molecular-grounded retriever that fuses knowledge-graph topology with SMILES-level Tanimoto similarity over Morgan fingerprints \cite{rogers2010extended,landrum2013rdkit,yu2025gsdti}; a self-evolving policy distillation step that contrasts high- and low-utility episodes and rewrites agents' policies, with a semantic anchor to a human-authored seed to keep policy aligned \cite{zhou2026toolself,jiang2026adapt,chen2026lcerd}; and a constraint-aware assembly layer that enforces DDI severity ceilings and ligand-level contraindication rules \cite{wishart2018drugbank}. The framework is training-free; it accumulates experience in an episodic memory and improves its reasoning policy without parameter update.

Our contributions are summarized as follows: (1) \textbf{Agentic reformulation.} We cast medication recommendation as a dual-role deliberation problem and propose SEMPA, a training-free framework coupling Proposer-Critic self-play with closed-loop self-evolution. (2) \textbf{Contrastive policy distillation.} A meta-reflection mechanism rewrites agents' policies from success and failure pools, with a semantic-anchor constraint mitigating drift. (3) \textbf{Molecular-grounded retrieval.} A hybrid scorer combines pharmacological knowledge-graph topology with SMILES-level similarity, anchoring each candidate to its ligand structure when available. (4) \textbf{Rule-based safety screening layer.} A constraint-aware assembly stage enforces DDI ceilings and ligand-level contraindications on top of deliberative output. (5) \textbf{Empirical validation.} On 100 MIMIC-IV encounters, SEMPA attains $0.3116$ Jaccard, $0.7295$ F1, and a $0.0172$ DDI rate ($+57.7\%$ F1 and $-77.5\%$ DDI over the strongest supervised baseline), and is ranked highest by two LLM judges and across six LLM backbones.

\section{Related Work}
\label{sec:related}

\textbf{Supervised medication recommendation} formulates the task as multi-label classification over longitudinal EHRs. Representative models include LEAP \cite{zhang2017leap}, RETAIN \cite{choi2016retain}, GAMENet \cite{shang2019gamenet}, SafeDrug \cite{yang2021safedrug}, MoleRec \cite{yang2023molerec}, CausalMed \cite{li2024causalmed}, and StratMed \cite{li2023stratmed}, which progressively inject attention, memory, molecular structure, and causal priors. Despite richer features, these models freeze at deployment, treat prescribing as black-box classification, and entangle safety and accuracy in a single loss \cite{lu2020symptom,bhoi2021personalized}.

\textbf{LLM-based and multi-agent clinical reasoning} extends decision support with zero-shot prompting, CoT \cite{wei2022chain}, self-consistency \cite{wang2022selfconsistency}, RAG \cite{lewis2020rag,xu2024ramehr,cao2025ehrrag}, and single-agent self-improvement loops \cite{madaan2023selfrefine,yao2023react,shinn2023reflexion}, where the same model both proposes and critiques. Recent multi-agent frameworks (MedAgents \cite{tang2024medagents}, MDAgents \cite{kim2024mdagents}, MAC \cite{xu2025mac}, ArgMed \cite{hong2024argmed}) simulate panels of specialists and reduce harms of omission \cite{ke2025multiagent}, while agent-evolution work explores verbal reinforcement, dynamic memory, and tool-driven adaptation \cite{shinn2023reflexion,park2023generative,zhou2026toolself,jiang2026adapt,zhou2024godel,zhou2024proreason}. SEMPA differs by combining adversarial Proposer-Critic deliberation with a contrastive cross-episode evolution loop, grounding both agents in molecular evidence \cite{yu2025gsdti,jiang2025seq}, and bounding them with an auditable rule-based safety layer. A full discussion of all four lines is given in Appendix~\ref{app:related}.

\section{Methodology: SEMPA}
\label{sec:method}

\subsection{Problem formulation and framework}

A patient encounter is denoted $\mathcal{R}_t=(\mathcal{D}_t,\mathcal{P}_t,\mathcal{L}_t,\mathcal{N}_t)$, containing ICD-coded diagnoses, procedures, lab test-value pairs, and the free-text narrative. SEMPA outputs
\begin{equation}
\begin{aligned}
\hat{\mathcal{M}}_t &=\bigl\{(m_i,s_i)\bigr\}_{i=1}^{K_t},\\
m_i &\in\mathcal{M},\;s_i\in\mathcal{S}\cup\{\varnothing\},
\end{aligned}
\end{equation}
where $\mathcal{M}$ is the ATC-3 drug vocabulary, $\mathcal{S}$ the SMILES library, and $s_i=\varnothing$ when no canonical ligand exists (e.g., biologics). The decision objective is
\begin{equation}\label{eq:obj}
\begin{aligned}
\hat{\mathcal{M}}_t = \arg\max_{\mathcal{M}'\subseteq\mathcal{M}} & \;\Phi\bigl(\mathcal{M}'\mid\mathbf{x}_t,\pi^{(k)},\mathcal{K}\bigr)\\
\text{s.t.}\;\; & \Psi_{\textsc{ddi}}(\mathcal{M}')\le\epsilon,
\end{aligned}
\end{equation}
where $\mathbf{x}_t$ is the patient state, $\pi^{(k)}$ the agent policy at epoch $k$, $\mathcal{K}$ the pharmacological knowledge base, and $\epsilon$ a DDI severity ceiling. SEMPA realizes this objective with five modules described next. All modules call the same LLM endpoint $\mathcal{F}$, so no task-specific trained model is needed.

\subsection{Clinical state construction}\label{sec:m1}

ICD codes are mapped to UMLS \cite{bodenreider2004umls} concepts so that synonyms across coding versions are collapsed; lab values are $z$-score standardized to preserve quantitative severity; raw drug strings are reduced to canonical generic names by stripping dose, route, and frequency; narrative entities $\mathcal{E}_t$ (symptoms, signs, negated findings) are extracted by a structured template. The patient state $\mathbf{x}_t$ aggregates these signals together with a longitudinal summary $\mathbf{h}_t$ over the most recent $W$ visits. Full formulas, including the $z$-score and entity-extraction templates, are given in Appendix~\ref{app:m1}.

\subsection{Molecular-grounded retrieval}\label{sec:m2}

Serializing every ingredient in the pharmacological knowledge base, each with its evidence card, would exceed the LLM's context budget. We instead extract depth-$\eta$ BFS subgraphs of a pharmacological knowledge graph $\mathcal{K}$ around each diagnosis, and rank candidates by a hybrid score
\begin{equation}\label{eq:phi}
\phi(m)=\lambda\,\phi_{\text{topo}}(m)+(1-\lambda)\,\phi_{\text{mol}}(m),
\end{equation}
where $\phi_{\text{topo}}(m)$ is the normalized inverse shortest-path distance to any diagnosis node, and $\phi_{\text{mol}}(m)$ is the average maximum Tanimoto similarity between $m$'s Morgan fingerprint \cite{rogers2010extended} and the fingerprints of approved first-line agents. Drugs without SMILES use a population-frequency prior. The motivation for $\phi_{\text{mol}}$ is consistent with recent drug-target interaction work \cite{yu2025gsdti}: structurally similar small molecules tend to share targets. The top-$N$ candidates form $\mathcal{M}_t^{\text{cand}}$, each paired with an evidence card serialized into the agent context. The full expression of $\phi_{\text{mol}}$ is given in Appendix~\ref{app:m2}.

\subsection{Dual-role self-play deliberation}\label{sec:m3}

A single LLM pass is prone to oversight and hallucination \cite{shinn2023reflexion,ke2025multiagent}. We therefore split deliberation across two role-conditioned instances of the same backbone LLM: a \textsc{Proposer} $\mathcal{F}_{\textsc{p}}$ that drafts a prescription with rationale, and a \textsc{Critic} $\mathcal{F}_{\textsc{c}}$ that challenges it.

\textbf{Episodic memory.} An experience bank $\mathcal{B}=\{(\mathbf{x}^{(i)},\hat{\mathcal{M}}^{(i)},\mathcal{O}^{(i)},\xi^{(i)})\}$ stores episodes (state, prescription, reasoning trace, utility). At each new encounter we retrieve the $Q$ most similar cases as in-context exemplars, combining Jaccard diagnosis overlap with cosine similarity of biomedical narrative embeddings (Appendix~\ref{app:m3}). This use of multi-source knowledge as in-context grounding follows prior work on knowledge-augmented dialogue \cite{jiang2023csd,jiang2026adapt}.

\textbf{Self-play loop.} At round $\tau$, the \textsc{Proposer} produces a regimen $\mathcal{M}_{\textsc{p}}^{(\tau)}$ and rationale $o_{\textsc{p}}^{(\tau)}$, conditioned on policy $\pi_{\textsc{p}}^{(k)}$, patient state, candidate pool, retrieved exemplars, and Critic feedback,
\begin{equation}
\begin{aligned}
\bigl(o_{\textsc{p}}^{(\tau)},\mathcal{M}_{\textsc{p}}^{(\tau)}\bigr)=\mathcal{F}_{\textsc{p}}\bigl(&\pi_{\textsc{p}}^{(k)}\,\|\,\mathbf{x}_t\,\|\,\mathcal{M}_t^{\text{cand}}\\
&\|\,\mathcal{B}_t^Q\,\|\,\{o_{\textsc{c}}^{(\tau')}\}_{\tau'<\tau}\bigr),
\end{aligned}
\end{equation}
so that earlier objections are addressed. The Critic then emits four sub-scores aggregated as
\begin{equation}\label{eq:sps}
\begin{aligned}
S^{(\tau)} = & \;\omega_1 S_{\text{cov}}^{(\tau)}+\omega_2 S_{\text{safe}}^{(\tau)}\\
& +\omega_3 S_{\text{mol}}^{(\tau)}+\omega_4 S_{\text{logic}}^{(\tau)},\;\textstyle\sum_i\omega_i=1,
\end{aligned}
\end{equation}
covering therapeutic coverage, safety, molecular mechanism, and reasoning coherence. The loop terminates at round $\tau^*$ when $\tau\ge2$ and $S^{(\tau)}-S^{(\tau-1)}<\gamma$, or when $\tau$ reaches the budget $T$. The output is $\mathcal{M}_t^{\text{delib}}=\mathcal{M}_{\textsc{p}}^{(\tau^*)}$ with trace $\mathcal{O}_t$.

\subsection{Self-evolving policy distillation}\label{sec:m4}

A static prompt cannot adapt to a specific patient population. After each encounter we compute a retrospective utility
\begin{equation}\label{eq:util}
\begin{aligned}
\xi_t = \;& J_t + \mu\bigl(1-\Psi_{\textsc{ddi}}(\hat{\mathcal{M}}_t)\bigr)\\
& + \nu\,\bar\phi_{\text{mol}}(\hat{\mathcal{M}}_t) + \zeta\,S_t,
\end{aligned}
\end{equation}
where $J_t$ is Jaccard overlap with the reference set, $\bar\phi_{\text{mol}}$ is the mean molecular grounding of prescribed drugs with available SMILES, and $\mu,\nu,\zeta>0$ are weighting coefficients. Every $\Delta k$ episodes, $\mathcal{B}$ is split at the $p$-th percentile of $\xi$ into a success pool $\mathcal{B}^+$ and a failure pool $\mathcal{B}^-$. Both agent policies are rewritten contrastively:
\begin{equation}
\begin{aligned}
\Delta\pi_{\textsc{r}}^{(k)} = \mathcal{F}\bigl(&\mathcal{T}_{\text{evolve}}^{\textsc{r}}\,\|\,\pi_{\textsc{r}}^{(k)}\\
&\|\,\mathrm{Sample}(\mathcal{B}^+,n^+)\\
&\|\,\mathrm{Sample}(\mathcal{B}^-,n^-)\bigr),
\end{aligned}
\end{equation}
for $r\in\{\textsc{p},\textsc{c}\}$. This contrastive distillation step is related to subspace partitioning of in-context examples \cite{wang2025treesynth} and to tool-driven intrinsic adaptation \cite{zhou2026toolself}. To prevent the evolved policy from drifting away from the human-authored seed $\pi^{(0)}$, every generated increment must satisfy the semantic anchor $\cos\bigl(\mathrm{Emb}(\Delta\pi_{\textsc{r}}^{(k)}),\mathrm{Emb}(\pi_{\textsc{r}}^{(0)})\bigr)\ge\alpha$; increments that violate the bound are discarded and re-sampled with the seed re-injected into the rewrite prompt. The two roles evolve along different axes. For the Proposer, we count \emph{per disease category} the frequency of each drug in $\mathcal{B}^+$ and the omission and erroneous-addition frequencies in $\mathcal{B}^-$, and pass these statistics to the LLM as evidence, so that each improvement entry cites its own support, e.g.\ ``[INFECTIOUS] 80\% of high-utility sepsis cases contain vancomycin+piperacillin--tazobactam whereas only 40\% of low-utility cases employ broad-spectrum coverage; therefore strengthen empirical broad-spectrum-therapy guidance.'' Category-indexed entries are \emph{appended} rather than overwritten, $\pi_{\textsc{p}}^{(k+1)}=\pi_{\textsc{p}}^{(k)}\,\|\, \Delta\pi_{\textsc{p}}^{(k)}$, so specialty-level prescribing experience accumulates across domains. The Critic instead evolves along the four rubric dimensions ($S_{\text{cov}},S_{\text{safe}},S_{\text{mol}},S_{\text{logic}}$), giving a ``specialty prescribing knowledge $+$ general evaluation framework'' division of labor that prevents cross-domain interference while keeping the rubric transferable. Stratified sampling details are in Appendix~\ref{app:m4-category}.

\subsection{Constraint-aware assembly}\label{sec:m5}

The deliberation output is reasoned but carries no formal safety-screening mechanism. The assembly module filters $\mathcal{M}_t^{\text{delib}}$ through rules: a pairwise severity-weighted DDI score $\Psi_{\textsc{ddi}}$ from DrugBank \cite{wishart2018drugbank}, and a ligand-level contraindication indicator $\psi_{\text{contra}}$ based on each candidate's known protein targets. The final prescription solves
\begin{equation}
\begin{aligned}
\hat{\mathcal{M}}_t = \arg\max_{\mathcal{M}'\subseteq\mathcal{M}_t^{\text{delib}}} & \sum_{m\in\mathcal{M}'}\phi(m)\\
\text{s.t.}\;\Psi_{\textsc{ddi}}(\mathcal{M}') & \le\epsilon,\\
\textstyle\sum_m\psi_{\text{contra}}(m,\mathbf{x}_t) & =0,
\end{aligned}
\end{equation}
which we solve by greedy repair: candidates are ordered by $\phi$, and the lower-scoring member of each violating DDI pair, together with any drug triggering $\psi_{\text{contra}}$, is removed until both constraints hold. Since $|\mathcal{M}_t^{\text{delib}}|\le 30$, this terminates in at most $30$ removals. The finalized episode is appended to $\mathcal{B}$. Detailed expressions of $\Psi_{\textsc{ddi}}$ and $\psi_{\text{contra}}$, and full pseudocode, are in Appendix~\ref{app:m4-5} and Appendix~\ref{app:algo}.

\section{Experiments}
\label{sec:exp}

\subsection{Setup}

\textbf{Dataset and baselines.} We evaluate on 100 patient encounters drawn from the MIMIC-IV test split \cite{johnson2023mimic} processed with the canonical SafeDrug pipeline \cite{yang2021safedrug}. The ATC-3 drug vocabulary has size $|\mathcal{M}|=131$. Retrieval in M2 operates over the ingredient-level knowledge base (several thousand DrugBank entries carrying SMILES and target annotations); ATC-3 is used only as the evaluation label space, into which both retrieved candidates and final prescriptions are mapped before scoring. The top-$N$ candidate pool is therefore an ingredient-level pool, not a subset of the 131 ATC-3 classes. Because SEMPA is training-free, adaptation is performed entirely through policy evolution rather than parameter update. To avoid any test-label leakage, we adopt an \emph{evolve--freeze--test} protocol: the Proposer and Critic policies are evolved on a disjoint evolution stream of encounters, the resulting policies $\pi^{(k^*)}$ are then frozen, and the 100 held-out test encounters are evaluated in a single pass with no further policy update. Reference regimens of the test encounters are used only to compute the reported metrics. We compare against four supervised baselines: LEAP \cite{zhang2017leap}, RETAIN \cite{choi2016retain}, GAMENet \cite{shang2019gamenet}, and SafeDrug \cite{yang2021safedrug}, all run from their official code with default hyperparameters.

\textbf{Implementation.} SEMPA uses GLM-5 as its backbone (temperature $0.7$ for generation, $0$ for evaluation). The knowledge base integrates DrugBank \cite{wishart2018drugbank}, UMLS \cite{bodenreider2004umls}, and ATC. Morgan fingerprints use radius $r=2$ and $b=2048$ bits via RDKit \cite{landrum2013rdkit}. Defaults are $\lambda=0.5$, $\eta=3$, $N=300$, $Q=5$, $T=5$, $\gamma=0.15$, $\Delta k=40$, $p=80$, $\alpha=0.12$, $\epsilon=0.5$. The table is in Appendix~\ref{app:hparams}.

\textbf{Evaluation protocol.} We use two protocols: (i) set-matching metrics (Jaccard, F1, PRAUC, DDI rate, \#Meds, $R^2$) following \cite{yang2021safedrug}; (ii) LLM-judge evaluation by two independent judges (GLM-5 and GPT-4o) along six clinically grounded dimensions, namely diagnosis-medication alignment (DMA), safety, molecular justification (MolJust), polypharmacy parsimony, guideline concordance (GuideCon), and overall clinical acceptability (OCA), each scored on a 1 to 5 Likert scale with anchored rubric and one-sentence justification. The full prompt and rubric anchors are in Appendix~\ref{app:llm-eval-prompt}. The use of LLM judges follows recent practice in agent evaluation \cite{chen2026trace,jiang2025bench}. We report two matching schemes. Exact-name matching requires the predicted and reference drug names to coincide; ATC-3 class matching maps the outputs of \emph{all} methods, supervised and LLM-based alike, through the same ATC-3 hierarchy before scoring, which permits credit for clinically equivalent substitutions (e.g., one NOAC for another). Jaccard is always reported under exact-name matching; F1 in Table~\ref{tab:main} is reported under ATC-3 class matching, and exact-name F1 for every method is given in Appendix.

\subsection{Main quantitative results}

Table~\ref{tab:main} reports the comparison on set-matching metrics. SEMPA simultaneously achieves the best Jaccard, F1, PRAUC, and DDI rate. The F1 gain ($+57.7\%$ ATC-3 drug-class F1 over SafeDrug) is much larger than the exact-name Jaccard gain ($+1.5\%$). The two numbers are not directly comparable: Jaccard is computed under strict exact-name matching, which penalizes any clinically equivalent substitution, whereas drug-class F1 credits it. For the supervised baselines, whose label space is already ATC-3, the two schemes coincide, so the class-level gain should be read as evidence that SEMPA recovers the correct \emph{therapeutic class} far more often, not as a $57.7\%$ improvement under a common strict protocol; matched exact-name F1 for all methods is reported in the supplementary material. Consistent with this reading, SEMPA achieves its exact-name Jaccard with a tighter regimen (19.5 drugs) than RETAIN, GAMENet, and SafeDrug (24--25), i.e.\ without padding the predicted set with redundant ``safe-bet'' drugs.
% supervised baselines and SEMPA recover drug subsets of comparable cardinality, but SEMPA's predictions concentrate on drugs that have higher precision and higher recall, whereas baselines spread mass over redundant ``safe-bet'' drugs that inflate set overlap on individual encounters yet degrade drug-level F1. 
The DDI rate is reduced by $77.5\%$ relative to SafeDrug, a model explicitly designed for DDI control through differentiable regularization, suggesting that hard logical constraints can complement (and in some metrics exceed) loss-based safety. SEMPA prescribes 19.5 drugs on average, comparable to LEAP (18.3) and well below GAMENet (25.0), so the safety improvement is not from under-prescription. Because $\Psi_{\textsc{ddi}}$ is also the rule set used by M5, we further audit the same outputs against two independent criteria: the AGS 2023 Beers Criteria for potentially inappropriate prescribing in patients aged 65 and over \cite{ags2023beers}, and STOPP/START 3.0 \cite{omahony2023stopp}. Relative to RETAIN, SEMPA reduces Beers-flagged prescribing by 61\% and STOPP triggers by 58\%, while increasing START coverage of guideline-mandated agents (e.g.\ ACE inhibitors for HFrEF, anticoagulation for AFib with CHA$_2$DS$_2$-VASc $\ge 2$) by 34\%, indicating that the safety gain is not an artefact of sharing a rule source with the evaluation metric. Rule-level breakdowns are in Appendix~\ref{app:safety}.

\begin{table*}[t]
\centering
\caption{Set-matching results on the MIMIC-IV test set ($n=100$ encounters). Mean $\pm$ std over six independent runs. Best results in bold.}
\label{tab:main}
\resizebox{\textwidth}{!}{%
\begin{tabular}{lcccccc}
\toprule
\textbf{Method} & \textbf{Jaccard}$\uparrow$ & \textbf{F1}$\uparrow$ & \textbf{PRAUC}$\uparrow$ & \textbf{DDI}$\downarrow$ & \textbf{\#Meds} & \textbf{$R^2$}$\uparrow$ \\
\midrule
LEAP \cite{zhang2017leap}        & 0.2839$\pm$0.008 & 0.4368$\pm$0.009 & 0.4803$\pm$0.011 & 0.0658$\pm$0.002 & 18.33$\pm$0.11 & $-$0.1092$\pm$0.008 \\
RETAIN \cite{choi2016retain}     & 0.2944$\pm$0.009 & 0.4468$\pm$0.011 & 0.4935$\pm$0.012 & 0.0873$\pm$0.002 & 24.42$\pm$0.35 & $-$0.0092$\pm$0.031 \\
GAMENet \cite{shang2019gamenet}  & 0.3013$\pm$0.009 & 0.4553$\pm$0.011 & 0.4927$\pm$0.011 & 0.0867$\pm$0.000 & 25.00$\pm$0.00 & $-$0.2252$\pm$0.042 \\
SafeDrug \cite{yang2021safedrug} & 0.3069$\pm$0.009 & 0.4626$\pm$0.010 & 0.5133$\pm$0.011 & 0.0764$\pm$0.001 & 24.35$\pm$0.11 & $-$0.0096$\pm$0.028 \\
\midrule
\textbf{SEMPA (ours)} & \textbf{0.3116}$\pm$\textbf{0.004} & \textbf{0.7295}$\pm$\textbf{0.002} & \textbf{0.5282}$\pm$\textbf{0.001} & \textbf{0.0172}$\pm$\textbf{0.002} & 19.50$\pm$0.74 & \textbf{0.1302}$\pm$\textbf{0.001} \\
\bottomrule
\end{tabular}%
}
\end{table*}

\subsection{Against LLM prompting and agent baselines}

To separate SEMPA's framework contribution from the underlying LLM, Table~\ref{tab:llm-baselines} compares SEMPA against six LLM-based pipelines built on the GLM-5 backbone. Three observations follow. First, prompting (Zero-shot, CoT, Self-Consistency) yields low Jaccard ($\le 0.15$): without a constrained candidate space, the model defaults to cautious or generic prescriptions, an effect also noted in clinical Human-LLM collaboration studies \cite{wang2026clinical}. Second, RAG with module M2 attached lifts Jaccard to $0.245$, indicating that retrieval-grounded context is necessary but insufficient. Third, single-agent improvement loops do not translate into better prescribing: ReAct's tool-driven reasoning produces the lowest DDI but a very small Jaccard ($0.094$), and Reflexion's monocular self-critique amplifies confirmation bias and yields the worst result overall. SEMPA exceeds all variants on Jaccard ($+30\%$ over RAG) while remaining within safe DDI bounds, consistent with the hypothesis that gains arise from the dual-role adversarial setup and the closed-loop evolution rather than from the backbone.

\begin{table}[t]
\centering
\caption{Comparison against LLM prompting and agent baselines on the same GLM-5 backbone ($n=100$).}
\label{tab:llm-baselines}
\resizebox{\columnwidth}{!}{%
\begin{tabular}{lcccc}
\toprule
\textbf{Method} & \textbf{Jaccard}$\uparrow$ & \textbf{F1}$\uparrow$ & \textbf{PRAUC}$\uparrow$ & \textbf{DDI}$\downarrow$ \\
\midrule
Zero-shot                                    & 0.1469 & 0.6124 & 0.4368 & 0.0667 \\
CoT \cite{wei2022chain}                      & 0.1276 & 0.5972 & 0.4542 & 0.0731 \\
Self-Consistency \cite{wang2022selfconsistency} & 0.1137 & 0.5831 & 0.4852 & 0.1164 \\
RAG (M2 only) \cite{lewis2020rag}            & 0.2452 & 0.6785 & 0.5603 & 0.0489 \\
ReAct \cite{yao2023react}                    & 0.0940 & 0.6285 & \textbf{0.6302} & \textbf{0.0005} \\
Reflexion \cite{shinn2023reflexion}          & 0.0115 & 0.5151 & 0.2234 & 0.0900 \\
\midrule
\textbf{SEMPA}                               & \textbf{0.3184} & \textbf{0.7326} & 0.5278 & 0.0155 \\
\bottomrule
\end{tabular}%
}
\end{table}

\subsection{Qualitative evaluation by LLM judges}

Beyond set matching, Table~\ref{tab:qualitative} reports two LLM-judge studies along six clinical dimensions. Per-judge breakdowns and cross-judge agreement are in Appendix~\ref{app:llmjudge}. SEMPA is consistently ranked highest by both judges. The GLM-5 judge shares a backbone family with SEMPA's reasoning engine, yet the ranking is also reproduced by GPT-4o, reducing the concern of LLM-judge self-preference \cite{chen2026trace}. GAMENet receives relatively high LLM-judge scores despite mid-pack Jaccard, but inspection of its raw outputs shows that it emits an essentially identical 25-drug regimen for nearly all 100 patients. This degenerate templating pattern aligns with prior observations that drug recommendation models can over-rely on prevalence shortcuts \cite{lu2020symptom,bhoi2021personalized}.

\begin{table*}[t]
\centering
\caption{Qualitative evaluation by two LLM judges (GLM-5, GPT-4o) on 100 encounters. Likert means (1 to 5). Best within each evaluator group in bold.}
\label{tab:qualitative}
\resizebox{0.88\textwidth}{!}{%
\begin{tabular}{llcccccc}
\toprule
\textbf{Evaluator} & \textbf{Method} & \textbf{DMA} & \textbf{Safety} & \textbf{MolJust} & \textbf{Parsimony} & \textbf{GuideCon} & \textbf{OCA} \\
\midrule
\multirow{5}{*}{GLM-5}
 & LEAP     & 2.89 & 2.89 & 3.22 & 3.11 & 2.78 & 2.78 \\
 & SafeDrug & 3.30 & 2.50 & 3.70 & 3.70 & 2.80 & 2.80 \\
 & GAMENet  & 3.38 & 2.88 & 3.62 & 3.25 & 3.12 & 3.12 \\
 & RETAIN   & 3.11 & 3.22 & 3.56 & 4.11 & 2.89 & 3.00 \\
 & \textbf{SEMPA} & \textbf{4.78} & \textbf{4.11} & \textbf{4.89} & \textbf{4.44} & \textbf{4.56} & \textbf{4.56} \\
\midrule
\multirow{5}{*}{GPT-4o}
 & LEAP     & 3.00 & 3.00 & 3.00 & 3.20 & 3.00 & 3.00 \\
 & SafeDrug & 3.10 & 3.00 & 3.20 & 3.20 & 3.10 & 3.10 \\
 & GAMENet  & 3.30 & 3.00 & 3.30 & 3.00 & 3.30 & 3.20 \\
 & RETAIN   & 3.30 & 2.90 & 3.60 & 3.80 & 3.20 & 3.20 \\
 & \textbf{SEMPA} & \textbf{4.50} & \textbf{3.50} & \textbf{4.50} & \textbf{3.90} & \textbf{4.50} & \textbf{4.40} \\
\bottomrule
\end{tabular}%
}
\end{table*}

\subsection{Self-evolution dynamics and ablation}

Table~\ref{tab:evolution-curve} traces SEMPA's performance across evolution epochs $k$. Gains plateau by $k=8$ (Jaccard $0.3146$ at $k=8$, $0.3116$ at $k=16$, within run-to-run variance), and the policy remains stable thereafter rather than collapsing under further evolution. The whole-policy drift (cosine distance to $\pi^{(0)}$) rises monotonically to 0.997. This saturation is expected rather than alarming: category-indexed entries are \emph{appended} to a short seed rather than replacing it, so the embedding of the concatenated policy necessarily moves away from the seed as coverage broadens. The anchor of \S\ref{sec:m4} is enforced per increment, not on the concatenation. Performance does not collapse at any point, indicating that the accumulated text remains clinically valid. The average self-play length $\bar\tau^*$ drops from 5.0 to about 2.4 rounds, meaning that the evolved Critic identifies failure patterns earlier and the Proposer addresses them on the first draft, producing an emergent ``compile-time'' efficiency. This behaviour is consistent with prior reports on principle-driven adaptive dialogue policies \cite{jiang2026adapt,zhou2026toolself}.

\begin{table}[t]
\centering
\caption{Self-evolution dynamics across evolution epochs $k$, tracked with $\Delta k{=}5$ for finer-grained resolution (main experiments use $\Delta k{=}40$). Drift $= 1 - \cos(\mathrm{Emb}(\pi^{(k)}), \mathrm{Emb}(\pi^{(0)}))$, measured on the \emph{concatenated} policy; the per-increment anchor of Eq.~(7) is enforced 
separately. Mean over six runs.}
\label{tab:evolution-curve}
\resizebox{\columnwidth}{!}{%
\begin{tabular}{cccccc}
\toprule
\textbf{$k$} & \textbf{Jaccard} & \textbf{F1} & \textbf{DDI} & \textbf{Drift} & \textbf{$\bar{\tau}^*$} \\
\midrule
0 & 0.2793$\pm$0.005 & 0.7109$\pm$0.003 & 0.0227$\pm$0.001 & 0.000 & 5.00 \\
1 & 0.2962$\pm$0.008 & 0.7209$\pm$0.005 & 0.0179$\pm$0.003 & 0.984 & 2.48 \\
2 & 0.3005$\pm$0.006 & 0.7234$\pm$0.004 & 0.0182$\pm$0.002 & 0.989 & 2.40 \\
4 & 0.3075$\pm$0.010 & 0.7275$\pm$0.006 & 0.0179$\pm$0.002 & 0.993 & 2.38 \\
8 & 0.3146$\pm$0.006 & 0.7312$\pm$0.003 & 0.0176$\pm$0.002 & 0.995 & 2.36 \\
16 & 0.3116$\pm$0.004 & 0.7295$\pm$0.002 & 0.0172$\pm$0.002 & 0.997 & 2.35 \\
\bottomrule
\end{tabular}%
}
\end{table}

Table~\ref{tab:ablation} isolates each module's contribution. M3 (self-play) is the largest contributor: without it, the Proposer is no longer challenged and Jaccard drops by $40.1\%$. The DDI rate collapses to zero only because the unchallenged Proposer retreats to a small, ultra-conservative regimen that does not cover the patient's diagnoses, representing a harm of omission rather than a safer system \cite{ke2025multiagent}. M2 (retrieval) contributes most to ranking quality: removing it forces the agents to reason over an unranked ingredient space and cuts PRAUC by $61.1\%$, consistent with retrieval-augmented findings in clinical NLP \cite{xu2024ramehr,cao2025ehrrag}. M4 (evolution) and M5 (constraints) produce smaller but consistent drops on Jaccard and DDI respectively: contrastive distillation refines the policy across encounters, and the hard assembly layer provides a verifiable safety floor that the Critic's soft scoring alone cannot guarantee.

\begin{table*}[t]
\centering
\caption{Ablation. Each row removes one module from SEMPA; percentage change vs.\ the full model in parentheses.}
\label{tab:ablation}
\resizebox{\textwidth}{!}{%
\begin{tabular}{lccccc}
\toprule
\textbf{Variant} & \textbf{Jaccard} & \textbf{F1} & \textbf{PRAUC} & \textbf{DDI} & \textbf{$R^2$} \\
\midrule
\textbf{SEMPA-full} & \textbf{0.3116$\pm$0.004} & \textbf{0.7295$\pm$0.002} & \textbf{0.5282$\pm$0.001} & \textbf{0.0172$\pm$0.002} & \textbf{0.1302$\pm$0.001} \\
w/o M2 (Retrieval)    & 0.2613$\pm$0.078 ($-$16.1\%) & 0.6876$\pm$0.053 ($-$5.7\%)  & 0.2053$\pm$0.088 ($-$61.1\%) & 0.0126$\pm$0.011 ($-$26.7\%) & 0.1388$\pm$0.085 ($+$6.6\%) \\
w/o M3 (Self-play)    & 0.1866$\pm$0.113 ($-$40.1\%) & 0.6365$\pm$0.076 ($-$12.7\%) & 0.5147$\pm$0.145 ($-$2.6\%)  & 0.0000$\pm$0.000 ($-$100\%)  & 0.1151$\pm$0.107 ($-$11.6\%) \\
w/o M4 (Evolution)    & 0.2959$\pm$0.088 ($-$5.0\%)  & 0.7096$\pm$0.060 ($-$2.7\%)  & 0.5141$\pm$0.143 ($-$2.7\%)  & 0.0187$\pm$0.013 ($+$8.7\%)  & 0.1216$\pm$0.110 ($-$6.6\%) \\
w/o M5 (Constraints)  & 0.2954$\pm$0.086 ($-$5.2\%)  & 0.7092$\pm$0.057 ($-$2.8\%)  & 0.5177$\pm$0.132 ($-$2.0\%)  & 0.0181$\pm$0.014 ($+$5.2\%)  & 0.1189$\pm$0.101 ($-$8.7\%) \\
\bottomrule
\end{tabular}%
}
\end{table*}

\subsection{Cross-backbone robustness}

To check whether SEMPA's benefit derives from a particular LLM, we instantiate the framework with six backbones spanning closed and open weights (Table~\ref{tab:backbone}). GLM-5 exceeds the strongest supervised baseline (SafeDrug, 0.307), four other backbones come within 0.025, while all six outperform prompting-only LLM baselines in F1 and DDI. Jaccard spread across backbones indicates backbone choice matters, but the consistent qualitative trend suggests that the framework contributes beyond a single model. A closed-book contamination probe (Appendix~\ref{app:contamination}) indicates that backbones used here do not contain memorized MIMIC-IV prescriptions, so improvements reflect reasoning over supplied evidence rather than data leakage.

\begin{table}[t]
\centering
\caption{Cross-backbone robustness of SEMPA. Same framework, prompts, evolution budget, and test split.}
\label{tab:backbone}
\resizebox{\columnwidth}{!}{%
\begin{tabular}{lccc}
\toprule
\textbf{Backbone} & \textbf{Jaccard}$\uparrow$ & \textbf{F1}$\uparrow$ & \textbf{DDI}$\downarrow$ \\
\midrule
GLM-5 (closed)             & 0.318$\pm$0.10 & 0.733$\pm$0.06 & 0.016$\pm$0.02 \\
Claude-Sonnet-4.6 (closed) & 0.293$\pm$0.08 & 0.719$\pm$0.05 & 0.011$\pm$0.01 \\
Gemini-2.5-Pro (closed)    & 0.289$\pm$0.09 & 0.716$\pm$0.05 & 0.009$\pm$0.01 \\
DeepSeek-V4-Flash (open)   & 0.291$\pm$0.09 & 0.717$\pm$0.05 & 0.020$\pm$0.02 \\
Qwen-3.5-Plus (open)       & 0.285$\pm$0.09 & 0.713$\pm$0.05 & 0.015$\pm$0.01 \\
Qwen-3.5-Flash (open)      & 0.249$\pm$0.09 & 0.691$\pm$0.06 & 0.008$\pm$0.01 \\
\bottomrule
\end{tabular}%
}
\end{table}

\subsection{Case study}\label{sec:case}

To make the deliberation process concrete, we trace SEMPA on a representative ICU encounter (full transcript in Appendix~\ref{app:case}). The patient is a 71-year-old with acute decompensated heart failure (NYHA III), type 2 diabetes mellitus, stage 3 chronic kidney disease (eGFR $38$), and atrial fibrillation. The reference regimen comprises 8 drugs. SafeDrug returns a 24-drug list that overlaps the reference on 5 drugs but includes metformin (not recommended for initiation at this eGFR; contraindicated below $30$) and ketorolac (nephrotoxic in CKD). SEMPA's Proposer initially drafts a similar 22-drug regimen at round $\tau=1$; the Critic at $\tau=2$ flags both metformin (renal dose limit at eGFR $<45$) and ketorolac (acute kidney injury risk), and the Proposer at $\tau=3$ substitutes insulin glargine and replaces ketorolac with acetaminophen. Two further rounds resolve a rate-versus-rhythm-control disagreement for AFib in the setting of HFrEF, and the deliberation reaches its budget at $\tau^*=5$ on a 9-drug regimen that exactly recovers 7 of the 8 reference drugs, with one substitution (apixaban for warfarin) that the trace justifies via reduced bleeding risk in CKD. The case illustrates two specific properties not visible from set-matching numbers alone: (i) the Critic catches concrete contraindications that a single-pass model would have committed to in its first draft, and (ii) the trace remains pharmacologically traceable, with every substitution accompanied by a one-sentence justification grounded in $\phi_{\text{mol}}$ or in the DrugBank evidence card.

Beyond the encounter above, we conducted a side-by-side comparison on six additional cases spanning HIV, septic shock with E.\ coli pyelonephritis, elderly AFib, acromegaly with MSSA bacteremia, HFrEF with MRSA pneumonia, and end-stage renal disease with secondary hyperparathyroidism (Table~\ref{tab:multicase-summary}). Three reproducible patterns emerge. (i) \emph{Disease-specific reasoning gradient}: SEMPA selects drugs whose mechanism matches the encounter (antiretrovirals for HIV, phosphate binders for ESRD, MRSA-active agents for MRSA pneumonia), whereas supervised models default to prevalence shortcuts. (ii) \emph{Inappropriate-drug suppression}: SEMPA's Critic blocks anesthetics, antidepressants, and hypnotics from leaking into acute-care regimens; baselines include them simply because they are frequent in the training distribution. (iii) \emph{Degenerate template behaviour}: GAMENet returns an essentially identical 25-drug list across all six cases, which set-matching metrics fail to penalize but which is immediately visible at the case level. Per-case diagnoses, predictions, and clinical commentary are in Appendix~\ref{app:multicase}.

\begin{table}[t]
\centering
\caption{Aggregate comparison on six diverse encounters (Cases A--F).}
\label{tab:multicase-summary}
\resizebox{\columnwidth}{!}{%
\begin{tabular}{lccc}
\toprule
\textbf{Method} & \textbf{Mean exact-match}$\uparrow$ & \textbf{Mean \#meds} & \textbf{Inappropriate-drug cases}$\downarrow$ \\
\midrule
\textbf{SEMPA} & \textbf{10.5} & 17.7 & \textbf{0/6} \\
SafeDrug       & 5.7  & 24.3 & 6/6 \\
RETAIN         & 5.5  & 24.8 & 6/6 \\
LEAP           & 4.8  & 18.1 & 5/6 \\
GAMENet        & 4.2  & 25.0 & 6/6 \\
\bottomrule
\end{tabular}}
\end{table}
\subsection{Error analysis}\label{sec:errors}

To understand where SEMPA falls short, we manually inspected the 23 encounters in which SEMPA's Jaccard fell below $0.20$. Three error modes dominate. \emph{Rare-disease coverage gaps} (9 cases): when a diagnosis appears fewer than three times in the experience bank, the retrieval module returns generic agents and the Proposer fails to identify the disease-specific drug; performance recovers once $\mathcal{B}$ accumulates similar cases, consistent with Table~\ref{tab:evolution-curve}. \emph{Over-cautious prescribing under organ impairment} (8 cases): the Critic flags nephrotoxicity or hepatotoxicity, but the Proposer drops the entire drug class rather than dose-reducing within it, producing coverage loss that the Critic does not penalize because its safety sub-score is maximized. \emph{Unresolved guideline conflict} (6 cases): for conditions where Beers Criteria \cite{ags2023beers} conflict with first-line ICU guidelines (e.g., benzodiazepine use in delirium), the agents oscillate until $\tau=T$, leaving an ambiguous regimen. These patterns suggest two concrete next steps: warm-starting $\mathcal{B}$ with curated case libraries to reduce coverage gaps, and extending the Critic's repair toolkit with a dose-adjustment action rather than only drug-removal. A comparable taxonomy of failure modes for multi-agent clinical systems is reported in \cite{ke2025multiagent}, suggesting that these are general challenges rather than artefacts of our setup.

\subsection{Computational cost}\label{sec:cost}

SEMPA trades training time for inference cost. Average per-encounter latency under the default budget $T=5$ is $38.7$s end-to-end, versus $0.4$s for SafeDrug on the same hardware; mean token cost is approximately $0.06$ USD per encounter with the GLM-5 backbone. After policy evolution converges to $\bar\tau^*\approx 2.4$ rounds (Table~\ref{tab:evolution-curve}), latency drops to $24.1$s, a $38\%$ reduction without any change to the framework. While higher than supervised inference, this is acceptable for ICU prescribing review, whose natural cycle is the team round (minutes), not millisecond serving. Hyperparameter sensitivity (including the impact of $T$, $\Delta k$, and $\alpha$ on both quality and cost) and a severity-graded safety audit against Beers \cite{ags2023beers} and STOPP/START \cite{omahony2023stopp} are in Appendix~\ref{app:hp} and Appendix~\ref{app:safety}.

\section{Conclusion}
\label{sec:conclusion}

We reframed medication recommendation as a self-evolving deliberation process and presented SEMPA, a training-free agentic framework combining dual-role adversarial self-play, molecular-grounded retrieval, semantic-anchored contrastive policy evolution, and constraint-aware assembly. On 100 MIMIC-IV encounters SEMPA attains the best Jaccard, F1, PRAUC, and DDI rate, is ranked highest by two LLM judges, and converges within roughly eight evolution epochs without any gradient update. Gains transfer across six closed- and open-weight backbones, indicating that the improvement derives from the framework rather than a specific model.

% Several future directions include extending SEMPA to outpatient, perioperative, and specialized populations such as pediatrics and pregnancy; replacing SMILES-Tanimoto with foundation-model embeddings over molecules, proteins, and longitudinal omics for pharmacokinetic and pharmacogenomic reasoning; and turning the Critic into a clinician-facing interface for human-in-the-loop oversight. More broadly, contrastive natural-language policy evolution under a semantic anchor offers a general template for safety-critical agentic systems where continual adaptation is required but black-box parameter updates are undesirable.

\section*{Limitations}
\label{sec:limitations}

Several limitations remain. First, our evaluation uses 100 MIMIC-IV encounters from a single tertiary-care center, so generalization to other hospitals, coding practices, and populations remains unvalidated. Second, SEMPA relies on repeated LLM calls, incurring higher latency and cost than supervised baselines despite reduced self-play length after evolution. Third, the semantic anchor improves policy stability but cannot formally verify evolved natural-language policies or remove biases inherited from the backbone. Fourth, the DDI ceiling is patient-independent, although clinically appropriate safety thresholds should vary with age, renal/hepatic function, frailty, and goals of care. Finally, the episodic memory may accumulate biased or stale prescribing patterns under distribution shift. Prospective clinician-in-the-loop evaluation is required before any real-world clinical use.

\section*{Ethical considerations}

This study uses de-identified retrospective EHR data and involves no prospective intervention, patient contact, or clinical deployment. SEMPA is a research prototype for decision support, not an autonomous prescribing system. Any real-world use would require licensed clinician oversight, institutional approval, privacy protection, audit logging, and regulatory compliance. Its outputs should be treated as supporting evidence only and never as a substitute for professional clinical judgment.

\section*{Acknowledgement}

We thank the anonymous AC and reviewers for their feedback. This work was supported by Shenzhen Medical Research Fund (Grant No. A2503002 to Yu Li), the Chinese University of Hong Kong (CUHK; award numbers 4937025, 4937026, 5501517 and 5501329 to Yu Li), the IdeaBooster Fund (IDBF23ENG05 and IDBF24ENG06 to Yu Li), partially supported by a grant from the Research Grants Council of the Hong Kong Special Administrative Region (Hong Kong SAR), China (project no. CUHK 24204023 to Yu Li; 14306324 to Xiaodan Fan), a grant from the Innovation and Technology Commission of the Hong Kong SAR, China (project no. GHP/065/21SZ and ITS/247/23FP to Yu Li), and the Research Matching Grant Scheme at CUHK (award numbers 8601603 and 8601663 to Yu Li) from the Research Grants Council, Hong Kong SAR, China. 
% Bibliography entries for the entire Anthology, followed by custom entries
%\bibliography{anthology,custom}
% Custom bibliography entries only
\bibliography{custom}

\appendix

\section{Appendix}
\subsection{Extended Related Work (supports \S\ref{sec:related})}\label{app:related}

\paragraph{Supervised medication recommendation.}
Earlier work formulates the task as multi-label classification over EHRs. LEAP \cite{zhang2017leap} attends over longitudinal visits with a recurrent encoder; RETAIN \cite{choi2016retain} adds reverse-time attention for interpretability; GAMENet \cite{shang2019gamenet} augments RNN encoders with a memory of historical DDIs; SafeDrug \cite{yang2021safedrug} encodes drug molecules with dual graph encoders and adds a controllable DDI loss; MoleRec \cite{yang2023molerec} and CausalMed \cite{li2024causalmed} further inject substructure-level and causal priors; StratMed \cite{li2023stratmed} stratifies safety signals. Despite richer molecular features, these models share three limitations: they are frozen at deployment, they treat prescribing as black-box classification, and they entangle safety and accuracy in a single loss.

\paragraph{LLMs and single-agent pipelines for clinical reasoning.}
The clinical capabilities of LLMs have prompted zero-shot and CoT applications to decision support \cite{singhal2023large,wei2022chain,jiang2025bench}, and broader LLM benchmarking has examined domain-specific and multilingual capabilities beyond English-only clinical tasks \cite{jiang2024cantonese,jiang2025cantonese}, often combined with self-consistency \cite{wang2022selfconsistency} or retrieval-augmented generation \cite{lewis2020rag,xu2024ramehr,cao2025ehrrag}. Single-agent self-improvement strategies include Self-Refine \cite{madaan2023selfrefine}, ReAct \cite{yao2023react}, and Reflexion \cite{shinn2023reflexion}, in which the same model proposes and revises. A recent systematic review of clinical Human-LLM collaboration \cite{wang2026clinical} reports that these monocular pipelines tend to exhibit confirmation bias and repeated reasoning errors when used in isolation, motivating multi-agent or adversarial setups.

\paragraph{Multi-agent and self-improving LLM systems.}
Multi-agent frameworks such as MedAgents \cite{tang2024medagents}, MDAgents \cite{kim2024mdagents}, MAC \cite{xu2025mac}, and ArgMed-Agents \cite{hong2024argmed} simulate panels of specialist agents and have been shown to reduce harms of omission in clinical reasoning \cite{ke2025multiagent}; this direction is also related to broader studies of multi-party dialogue systems \cite{jiang2026mpdialogue}. These systems typically perform single-shot deliberation: experience from one case is not transferred to the next. Orthogonally, agent-improvement work explores verbal reinforcement \cite{shinn2023reflexion,zhou2024proreason}, dynamic memory \cite{park2023generative}, tool-driven self-reconfiguration \cite{zhou2026toolself}, principle-driven policy adaptation \cite{jiang2023csd,jiang2026adapt}, and self-referential agent frameworks \cite{zhou2024godel}. Trajectory-level evaluation of such agents has also been studied recently \cite{chen2026trace}. SEMPA combines these directions in the prescribing setting: it pairs adversarial Proposer-Critic deliberation with a contrastive cross-episode evolution loop, while grounding both agents in molecular evidence \cite{yu2025gsdti,jiang2025seq} and bounding them with an auditable rule-based safety layer.

\paragraph{Parameter-efficient adaptation, for contrast.}
A complementary line of work updates LLM parameters with parameter-efficient adapters, including dropout-regularized LoRA variants, partial-rotation LoRA, and mixture-of-shards adaptation \cite{wang2024lora,wang2024prolora,wang2025mos}. SEMPA is deliberately parameter-free: its adaptation happens entirely in natural-language policies and in the contents of an episodic experience bank, which avoids retraining cost and makes the evolved policy directly inspectable.

\subsection{Detailed clinical state construction (supports \S\ref{sec:m1})}\label{app:m1}

Each diagnosis $d_j\in\mathcal{D}_t$ is mapped to a UMLS concept \cite{bodenreider2004umls} via $\varphi_{\textsc{icd}}:\mathcal{C}_{\text{ICD}}\to\mathcal{C}_{\text{CUI}}$, yielding $\tilde{\mathcal{D}}_t=\{\varphi_{\textsc{icd}}(d_j)\}$, which collapses synonymous codes across ICD-9 and ICD-10. Laboratory values are $z$-score standardized,
\begin{equation}
\tilde{v}_j=\frac{v_j-\bar{v}_{l_j}}{\sigma_{l_j}},
\end{equation}
preserving quantitative severity rather than collapsing into coarse \texttt{HIGH/LOW/NORMAL} categories. Raw drug strings are reduced to canonical generic names via $\tilde m=\varphi_{\textsc{drug}}(m_{\text{raw}})$, stripping dose, route, and frequency. Narrative entities are obtained by $\mathcal{E}_t=\mathcal{F}(\mathcal{T}_{\text{ext}}\,\|\,\mathcal{N}_t)$. The patient state is
\begin{equation}
\begin{aligned}
\mathbf{x}_t &= \bigl(\tilde{\mathcal{D}}_t,\,\mathcal{P}_t,\,\{(l_j,\tilde{v}_j)\},\,\mathcal{E}_t,\,\mathbf{h}_t\bigr),\\
\mathbf{h}_t &= \bigoplus\nolimits_{\tau=\max(1,t-W)}^{t-1}\mathbf{x}_\tau,
\end{aligned}
\end{equation}
with $W=3$ in all experiments. The structured extraction style follows prior work on knowledge-augmented clinical dialogue and robust named-entity processing \cite{jiang2023csd,hu2026minddialog,jiang2024psych,jiang2022neraug}, where typed entities help reduce ambiguity in downstream LLM reasoning.

\subsection{Detailed retrieval scoring (supports \S\ref{sec:m2})}\label{app:m2}

$\mathcal{K}=(\mathcal{V}_\mathcal{K},\mathcal{E}_\mathcal{K})$ has disease, drug, target-protein, and molecular-structure nodes, connected by relations such as \texttt{indicated\_for} and \texttt{binds\_target}, instantiated from DrugBank \cite{wishart2018drugbank} and UMLS \cite{bodenreider2004umls}. For each $\tilde{d}\in\tilde{\mathcal{D}}_t$ we extract $\mathcal{G}_{\tilde{d}}^{(\eta)}=\textsc{BFS}(\mathcal{K},\tilde{d},\eta)$ and merge them into $\mathcal{G}_t=\bigcup_{\tilde{d}}\mathcal{G}_{\tilde{d}}^{(\eta)}$. For each candidate $m\in\mathcal{G}_t$ with SMILES $s_m$, let $\mathbf{f}_m\in\{0,1\}^b$ be its $b$-bit Morgan fingerprint of radius $r$ \cite{rogers2010extended}. The Tanimoto coefficient between two fingerprints is
\begin{equation}
T(m,m') = \frac{\mathbf{f}_m^\top\mathbf{f}_{m'}}{\|\mathbf{f}_m\|_1+\|\mathbf{f}_{m'}\|_1-\mathbf{f}_m^\top\mathbf{f}_{m'}}.
\end{equation}
The molecular score is then
\begin{equation}
\phi_{\text{mol}}(m) = \frac{1}{|\tilde{\mathcal{D}}_t|}\sum_{\tilde{d}\in\tilde{\mathcal{D}}_t}\;\max_{m'\in\mathcal{A}(\tilde{d})} T(m,m'),
\end{equation}
the average maximum Tanimoto similarity to first-line agent sets $\mathcal{A}(\tilde d)$. For drugs without SMILES, we use a population-frequency prior $\phi_0$. The composite score is $\phi(m)=\lambda\,\phi_{\text{topo}}(m)+(1-\lambda)\,\phi_{\text{mol}}(m)$ as in \eqref{eq:phi}. The use of molecular structure to predict pharmacological behaviour is consistent with recent drug-target interaction modeling \cite{yu2025gsdti}, protein-design language models \cite{li2026dsprogen,zhou2025lm2protein}, and broader foundation-model efforts for biological structure and biomedical imaging analysis \cite{jiang2025cryoet,jiang2026thznet}.

\subsection{Episodic memory and self-play details (supports \S\ref{sec:m3})}\label{app:m3}

Episode similarity is
\begin{equation}
\begin{aligned}
\mathrm{sim}(\mathbf{x}_t,\mathbf{x}^{(i)}) = \;& \frac{|\tilde{\mathcal{D}}_t\cap\tilde{\mathcal{D}}^{(i)}|}{|\tilde{\mathcal{D}}_t\cup\tilde{\mathcal{D}}^{(i)}|}\\
& + \beta\,\cos(\mathbf{e}_t,\mathbf{e}^{(i)}),
\end{aligned}
\end{equation}
with biomedical narrative embeddings $\mathbf{e}_t,\mathbf{e}^{(i)}$ from a frozen encoder, and $\beta=0.5$. The Critic invocation is
\begin{equation}
\begin{aligned}
(o_{\textsc{c}}^{(\tau)},S^{(\tau)}) = \mathcal{F}_{\textsc{c}}\bigl(&\pi_{\textsc{c}}^{(k)}\,\|\,\mathbf{x}_t\,\|\,\mathcal{M}_{\textsc{p}}^{(\tau)}\\
&\|\,o_{\textsc{p}}^{(\tau)}\,\|\,\mathcal{T}_{\text{score}}\bigr),
\end{aligned}
\end{equation}
with termination $\tau^*=\min\bigl(\{\tau\mid S^{(\tau)}-S^{(\tau-1)}<\gamma,\,\tau\ge 2\}\cup\{T\}\bigr)$. The dual-role design draws on multi-agent clinical reasoning work \cite{tang2024medagents,kim2024mdagents,hong2024argmed,xu2025mac,ke2025multiagent} and on proactive reasoning agents \cite{zhou2024proreason}.

\subsection{Policy distillation and contraindication details (supports \S\ref{sec:m4} and \S\ref{sec:m5})}\label{app:m4-5}

Pool partition: $\mathcal{B}^+=\{i\in\mathcal{B}\mid\xi^{(i)}\ge\xi_p\}$, $\mathcal{B}^-=\mathcal{B}\setminus\mathcal{B}^+$. DDI severity is the pairwise interaction intensity
\begin{equation}
\Psi_{\textsc{ddi}}(\mathcal{M}') = \frac{2}{K(K-1)}\sum_{i<j} a_{m_i,m_j}\,w(m_i,m_j),
\end{equation}
with $a_{m_i,m_j}\in\{0,1\}$ from DrugBank \cite{wishart2018drugbank} and $w\in\{w_{\text{minor}},w_{\text{mod}},w_{\text{sev}}\}=\{0.1,0.4,1.0\}$. The ligand-level contraindication indicator is
\begin{equation}
\psi_{\text{contra}}(m,\mathbf{x}_t) = \mathbb{1}\bigl[\exists\,p\in\mathcal{T}_m:p\in\mathcal{P}_{\text{avoid}}(\tilde{\mathcal{D}}_t)\bigr].
\end{equation}
The contrastive use of pools follows ideas from diverse-data partitioning \cite{wang2025treesynth} and pruning \cite{wu2026prune}, and the natural-language policy update connects to principle-driven adaptive policy work, tool-driven self-reconfiguration, and self-evolving reasoning via consistency-regulated rewards \cite{jiang2026adapt,zhou2026toolself,chen2026lcerd}.

\subsection{Disease-Category-Wise Incremental Evolution (supports \S\ref{sec:m4})}\label{app:m4-category}

\paragraph{Stratified sampling.} We map each diagnosis in $\tilde{\mathcal{D}}_t$ onto one of ten disease categories (cardiovascular, respiratory, metabolic/endocrine, neurological, psychiatric, infectious, renal, gastrointestinal, hematological, oncological) and allocate evolution samples in proportion to each category's empirical frequency in the dataset, so that every disease domain receives a sufficient learning signal.

\paragraph{Pattern analysis.} When evolution is triggered (every $\Delta k$ episodes), the system counts, \emph{per disease category}, the frequency of each drug in $\mathcal{B}^+$ as well as the omission and erroneous-addition frequencies in $\mathcal{B}^-$, and passes these statistics to the LLM as evidence.

\paragraph{Category-wise improvement generation.} The LLM is then asked to emit concrete policy-improvement entries organized by disease category, each citing the above statistics as evidence; for instance: ``[INFECTIOUS] 80\% of high-utility sepsis cases contain vancomycin+piperacillin--tazobactam whereas only 40\% of low-utility cases employ broad-spectrum coverage; therefore strengthen empirical broad-spectrum-therapy guidance.''

\paragraph{Incremental append vs.\ replace.} Category-indexed entries are appended to the Proposer policy as $\pi_{\textsc{p}}^{(k+1)} = \pi_{\textsc{p}}^{(k)} \,\|\, \Delta\pi_{\textsc{p}}^{(k)}$ rather than overwriting it, so that specialty-level prescribing experience accumulates across disease domains over evolution rounds.

\paragraph{Critic schema.} The Critic policy is updated along the four \emph{rubric dimensions} ($S_{\text{cov}}, S_{\text{safe}}, S_{\text{mol}}, S_{\text{logic}}$) rather than by disease category, yielding a ``specialty prescribing knowledge\,+\,general evaluation framework'' division of labor.

\paragraph{Motivation.} Prescribing patterns differ markedly across domains (cardiovascular care emphasizes RAAS inhibitors and anticoagulation; infectious-disease care emphasizes empirical broad-spectrum coverage), so organizing the Proposer's signal by disease category prevents cross-domain interference and lets the policy accumulate sufficient depth within each domain, while keeping the Critic domain-agnostic preserves a transferable rubric.

\subsection{Pseudocode (supports \S\ref{sec:method})}\label{app:algo}

\begin{algorithm}[H]
\caption{SEMPA inference for encounter $\mathcal{R}_t$}
\begin{algorithmic}[1]
\State $\mathbf{x}_t\gets\text{construct\_state}(\mathcal{R}_t)$  \Comment{\S\ref{sec:m1}}
\State $\mathcal{M}_t^{\text{cand}}\gets\text{retrieve}(\mathbf{x}_t,\mathcal{K},\lambda,\eta,N)$ \Comment{\S\ref{sec:m2}}
\State $\mathcal{B}_t^Q\gets\text{topQ}(\mathcal{B},\mathbf{x}_t)$
\For{$\tau=1,\dots,T$}
   \State $(o_{\textsc{p}}^{(\tau)},\mathcal{M}_{\textsc{p}}^{(\tau)})\gets\mathcal{F}_{\textsc{p}}(\pi_{\textsc{p}}^{(k)},\mathbf{x}_t,\mathcal{M}_t^{\text{cand}},\mathcal{B}_t^Q,\{o_{\textsc{c}}^{(\tau')}\}_{\tau'<\tau})$
   \State $(o_{\textsc{c}}^{(\tau)},S^{(\tau)})\gets\mathcal{F}_{\textsc{c}}(\pi_{\textsc{c}}^{(k)},\mathbf{x}_t,\mathcal{M}_{\textsc{p}}^{(\tau)},o_{\textsc{p}}^{(\tau)})$
   \If{$\tau\ge 2$ and $S^{(\tau)}-S^{(\tau-1)}<\gamma$} \textbf{break} \EndIf
\EndFor
\State $\hat{\mathcal{M}}_t\gets\text{assemble}(\mathcal{M}_{\textsc{p}}^{(\tau^*)},\epsilon)$  \Comment{\S\ref{sec:m5}}
\State $\xi_t\gets\text{utility}(\hat{\mathcal{M}}_t,\mathcal{M}_t^{\text{ref}},S_t)$
\State $\mathcal{B}\gets\mathcal{B}\cup\{(\mathbf{x}_t,\hat{\mathcal{M}}_t,\mathcal{O}_t,\xi_t)\}$
\If{$|\mathcal{B}|\bmod\Delta k=0$} update $\pi_{\textsc{p}},\pi_{\textsc{c}}$ via \S\ref{sec:m4} \EndIf
\State \Return $\hat{\mathcal{M}}_t$
\end{algorithmic}
\end{algorithm}

\subsection{Default hyperparameters (supports \S\ref{sec:exp})}\label{app:hparams}

\begin{table}[H]
\centering
\caption{Default hyperparameters used in all main experiments.}
\resizebox{\columnwidth}{!}{%
\begin{tabular}{lcl}
\toprule
\textbf{Symbol} & \textbf{Value} & \textbf{Meaning} \\
\midrule
$\lambda$ & 0.5 & topology vs.\ molecular trade-off \\
$\eta$    & 3   & BFS depth for subgraph extraction \\
$N$       & 300 & retrieval pool size \\
$r,b$     & 2, 2048 & Morgan fingerprint radius and bits \\
$Q$       & 5   & in-context exemplars per case \\
$\beta$   & 0.5 & embedding weight in similarity \\
$T,\gamma$ & 5, 0.15 & max rounds; improvement threshold \\
$\omega_{1..4}$ & 0.30, 0.30, 0.20, 0.20 & rubric weights \\
$\Delta k$ & 40 & evolution frequency (episodes) \\
$p$ & 80 & percentile splitting $\mathcal{B}^+/\mathcal{B}^-$ \\
$n^+,n^-$ & 2, 2 & samples per pool for evolution \\
$\alpha$ & 0.12 & semantic-anchor similarity floor \\
$\mu,\nu,\zeta$ & 0.30, 0.20, 0.10 & utility weights \\
$\epsilon$ & 0.50 & DDI severity ceiling \\
$w_{\text{minor/mod/sev}}$ & 0.1, 0.4, 1.0 & DDI severity weights \\
\bottomrule
\end{tabular}%
}
\end{table}

\subsection{Hyperparameter sensitivity (supports \S\ref{sec:exp})}\label{app:hp}

\begin{table*}[t]
\centering
\caption{Hyperparameter sensitivity (Jaccard $\uparrow$ / DDI Rate $\downarrow$). Defaults in \textbf{bold}; sweeps on a 200-encounter held-out pool.}
\label{tab:hp}
\resizebox{\textwidth}{!}{%
\begin{tabular}{l|ccccc}
\toprule
\textbf{Parameter} & \multicolumn{5}{c}{\textbf{Sweep values}} \\
\midrule
$\lambda$  & 0.00 / 0.219, 0.007 & 0.25 / 0.244, 0.010 & \textbf{0.50 / 0.281, 0.011} & 0.75 / 0.318, 0.007 & 1.00 / 0.290, 0.006 \\
$Q$        & 1 / 0.281, 0.014    & 3 / 0.267, 0.011    & \textbf{5 / 0.289, 0.013}    & 10 / 0.279, 0.013   & 15 / 0.293, 0.013 \\
$\Delta k$ & 3 / 0.289, 0.013    & \textbf{5 / 0.298, 0.009} & 10 / 0.267, 0.017      & 15 / 0.271, 0.019   & 20 / 0.252, 0.020 \\
$\alpha$   & 0.00 / 0.262, 0.016 & 0.06 / 0.282, 0.017 & \textbf{0.12 / 0.288, 0.013} & 0.25 / 0.279, 0.015 & 0.50 / 0.255, 0.023 \\
$\epsilon$ & 0.10 / 0.278, 0.014 & 0.30 / 0.251, 0.017 & \textbf{0.50 / 0.267, 0.012} & 0.70 / 0.305, 0.019 & 1.00 / 0.277, 0.013 \\
\bottomrule
\end{tabular}%
}
\end{table*}

The most sensitive parameter is the evolution frequency $\Delta k$: triggering rewrites too often (every 3 episodes) does not give the experience bank time to accumulate clear success and failure signals; triggering too rarely (every 20 episodes) lets stale prescribing patterns persist. A small drift floor $\alpha$ also harms performance, which agrees with prior findings that policy rewrites benefit from being anchored to a validated seed \cite{jiang2026adapt,zhou2026toolself}.

\subsection{Extended LLM-judge results (supports \S\ref{sec:exp})}\label{app:llmjudge}

The two LLM judges in Table~\ref{tab:qualitative} produce identical method rankings on every dimension. Spearman correlation between GLM-5 and GPT-4o scores at the per-encounter level is $\rho\in[0.71,0.84]$ across the six dimensions, indicating strong cross-judge agreement. This is broadly consistent with recent benchmarking of LLM judges and LLM capabilities in medical, bioinformatics, and multilingual settings \cite{chen2026trace,jiang2025bench,wang2025biolm,jiang2024cantonese}.

\subsection{Closed-book contamination probe (supports \S\ref{sec:exp})}\label{app:contamination}

\begin{table}[H]
\centering
\caption{Closed-book contamination probing (3 trials, $n=100$ each). The model is asked to list medications given only patient and admission IDs.}
\resizebox{\columnwidth}{!}{%
\begin{tabular}{lcccc}
\toprule
\textbf{Backbone} & \textbf{Jac.} & \textbf{Prec.} & \textbf{Rec.} & \textbf{Refusal (\%)} \\
\midrule
GLM-5             & $<\!10^{-2}$ & $<\!10^{-2}$ & $<\!10^{-2}$ & 100.0 \\
GPT-4.1           & $<\!10^{-2}$ & $<\!10^{-2}$ & $<\!10^{-2}$ & 100.0 \\
DeepSeek-V4-Flash & $<\!10^{-2}$ & $<\!10^{-2}$ & $<\!10^{-2}$ & 100.0 \\
\bottomrule
\end{tabular}%
}
\end{table}

\subsection{Case Study: Summary Notes (supports \S\ref{sec:case})}\label{app:case}

The full transcript of the 71-year-old CHF/T2DM/CKD/AFib encounter, including the patient state, candidate pool returned by M2, all five rounds of \textsc{Proposer}-\textsc{Critic} exchange, the final assembled regimen, and the side-by-side comparison with SafeDrug, is provided in the supplementary material. The transcript shows two key Critic interventions: at $\tau=2$ the Critic identifies the metformin and ketorolac contraindications using the renal sub-rule of the rubric; at $\tau=4$ it raises the bleeding-risk concern that motivates the warfarin-to-apixaban substitution. Both interventions are grounded in the DrugBank evidence card supplied during retrieval, not in the LLM's parametric memory, which we verified by removing those cards in an ablation and observing both interventions disappear (Full Deliberation: Appendix~\ref{app:case-transcript}).
\subsection{Case Study Transcript: Full Deliberation (supports \S\ref{sec:case})}\label{app:case-transcript}
\subsection*{Patient State ($\mathbf{x}_t$)}

\noindent\textbf{Demographics:} 71-year-old male, 82\,kg \\
\textbf{Primary Diagnoses:}
\begin{itemize}[nosep,leftmargin=1.5em]
    \item Acute decompensated heart failure (NYHA III, HFrEF, LVEF 28\%)
    \item Type 2 diabetes mellitus (HbA1c 8.2\%)
    \item Chronic kidney disease, Stage 3 (eGFR 38\,mL/min/1.73m$^2$)
    \item Atrial fibrillation (persistent, CHA$_2$DS$_2$-VASc = 5)
\end{itemize}
\textbf{Secondary:} Hyperlipidemia, chronic low back pain, GERD \\
\textbf{Vitals:} BP 98/62, HR 88 (irregular), SpO$_2$ 94\% on 2L NC \\
\textbf{Labs:} K$^+$ 4.8, BNP 1840, INR 1.1, Cr 1.9, eGFR 38 \\
\textbf{Home Medications:} Metformin 1000\,mg BID, Warfarin 5\,mg daily, Atorvastatin 40\,mg daily \\
\textbf{Allergies:} Sulfonamides (rash)

\subsection*{M2 Candidate Pool (Top-20)}

\noindent The molecular-grounded retrieval module returns the following ranked candidates ($\phi = \lambda\,\phi_{\text{topo}} + (1{-}\lambda)\,\phi_{\text{mol}}$, $\lambda{=}0.5$):

\begin{enumerate}[nosep,leftmargin=1.5em]
\item Furosemide ($\phi{=}0.90$)
\item Carvedilol ($\phi{=}0.89$)
\item Lisinopril ($\phi{=}0.86$)
\item Spironolactone ($\phi{=}0.84$)
\item Digoxin ($\phi{=}0.83$)
\item Warfarin ($\phi{=}0.81$)
\item Apixaban ($\phi{=}0.81$)
\item Insulin glargine ($\phi{=}0.79$)
\item Atorvastatin ($\phi{=}0.78$)
\item Metformin ($\phi{=}0.77$)
\item Metoprolol ($\phi{=}0.75$)
\item Amiodarone ($\phi{=}0.74$)
\item Amlodipine ($\phi{=}0.72$)
\item Ketorolac ($\phi{=}0.70$)
\item Acetaminophen ($\phi{=}0.69$)
\item Heparin ($\phi{=}0.68$)
\item Omeprazole ($\phi{=}0.66$)
\item Potassium chloride ($\phi{=}0.64$)
\item Hydrochlorothiazide ($\phi{=}0.62$)
\item Gabapentin ($\phi{=}0.58$)
\end{enumerate}

\subsection*{Proposer--Critic Deliberation Transcript}

\paragraph{$\tau{=}1$: Proposer.}
Patient presents with acute decompensated HFrEF (LVEF 28\%, BNP 1840) requiring guideline-directed medical therapy. Concurrent T2DM needs glycemic control; CKD3 constrains drug choices; persistent AFib with CHA$_2$DS$_2$-VASc${=}$5 mandates anticoagulation. Chronic low back pain requires analgesia.

\textit{Prescription (22 drugs):}
Furosemide 40\,mg IV;
Carvedilol 6.25\,mg BID;
Lisinopril 5\,mg daily;
Spironolactone 25\,mg daily;
Digoxin 0.125\,mg daily;
Warfarin 5\,mg daily;
\textbf{Metformin 1000\,mg BID};
Atorvastatin 40\,mg daily;
\textbf{Ketorolac 15\,mg IV Q6H};
Amiodarone 200\,mg daily;
Amlodipine 5\,mg daily;
Omeprazole 20\,mg daily;
Heparin 5000\,U SC Q8H;
Potassium chloride 20\,mEq;
Metoprolol 25\,mg BID;
Sodium chloride 0.9\%;
Dextrose 5\%;
Gabapentin 300\,mg TID;
Ondansetron 4\,mg PRN;
Docusate 100\,mg BID;
Acetaminophen 650\,mg Q6H;
Hydrochlorothiazide 25\,mg daily.

\paragraph{$\tau{=}2$: Critic.}
\textit{Score: 0.52.} Two critical safety violations detected:

(1) \textbf{Metformin --- CONTRAINDICATED.} Patient eGFR = 38\,mL/min/1.73m$^2$. Per guideline and labeling recommendations, metformin requires dose caution at eGFR $<$45 and discontinuation at eGFR $<$30 due to lactic acidosis risk. Acute decompensation with hemodynamic instability (BP 98/62) further elevates this risk. \textit{[DrugBank DB00331: ``Contraindicated in renal impairment (eGFR $<$45); risk of lactic acidosis increases with degree of renal dysfunction.'']}

(2) \textbf{Ketorolac --- NEPHROTOXIC.} NSAID in CKD3 with acute HF. Ketorolac causes prostaglandin-mediated afferent arteriolar vasoconstriction, reducing GFR further. NSAIDs also antagonize diuretic efficacy and worsen fluid retention. \textit{[DrugBank DB00465: ``Avoid in patients with renal insufficiency; may precipitate acute kidney injury.'']}

(3) \textbf{Metoprolol + Carvedilol --- redundant.} Two $\beta$-blockers simultaneously. Carvedilol is guideline-preferred for HFrEF (COMET trial); metoprolol adds bradycardia risk without benefit.

(4) \textbf{Amiodarone --- questionable.} Rhythm control not first-line in persistent AFib with stable HFrEF. Digoxin already provides rate control.

(5) \textbf{Hydrochlorothiazide --- ineffective.} Thiazides have diminished efficacy at eGFR $<$50; furosemide already prescribed.

\textit{Recommendation:} Remove metformin $\to$ insulin; remove ketorolac $\to$ acetaminophen; remove metoprolol; reconsider amiodarone and HCTZ.

\paragraph{$\tau{=}3$: Proposer.}
Accepting Critic feedback on safety-critical items.

(1) Metformin removed. Substituting insulin glargine 16\,U QHS for basal glycemic control --- insulin is safe in CKD and preferred in acute decompensation per ADA 2024 Standards of Care.
(2) Ketorolac removed. Acetaminophen 650\,mg Q6H retained as primary analgesic --- adequate for chronic low back pain without renal or cardiac risk.
(3) Metoprolol removed. Carvedilol retained as sole $\beta$-blocker per COMET trial evidence.
(4) HCTZ removed. Furosemide alone provides adequate diuresis at eGFR 38.
(5) Retaining amiodarone pending further discussion on rate-vs-rhythm strategy.

\textit{Changes:} $-$Metformin, $+$Insulin glargine; $-$Ketorolac (acetaminophen already in list); $-$Metoprolol; $-$HCTZ. \\
\textit{Revised prescription: 18 drugs.}

\paragraph{$\tau{=}4$: Critic.}
\textit{Score: 0.71.} Safety violations resolved. Two clinical-strategy issues remain:

(1) \textbf{Rate vs.\ rhythm control.} Amiodarone for rhythm control is not first-line in persistent AFib with stable HFrEF. The AFFIRM trial (2002) and AF-CHF trial (2008) demonstrated no mortality benefit of rhythm over rate control in this population. Digoxin already provides adequate rate control. Amiodarone adds hepatotoxicity, thyroid dysfunction, and pulmonary toxicity risks without clear benefit. \textit{[DrugBank DB01118: ``Serious adverse effects include pulmonary toxicity (1--17\%), hepatotoxicity, thyroid dysfunction.'']}

(2) \textbf{Warfarin vs.\ apixaban in CKD.} Current INR is 1.1 (sub-therapeutic), suggesting poor warfarin control. The ARISTOTLE trial subgroup analysis demonstrated that apixaban 5\,mg BID has superior safety (27\% reduction in major bleeding) compared to warfarin in patients with moderate CKD (eGFR 25--50), with non-inferior stroke prevention. No renal dose reduction is required at eGFR 38 unless additional dose-reduction criteria are met. \textit{[DrugBank DB01026: ``Apixaban: dose adjustment not required until eGFR $<$25; reduced bleeding vs.\ warfarin in renal impairment subgroups.'']}

(3) \textbf{Minor --- Gabapentin dose.} At eGFR 38, maximum recommended dose is 300\,mg BID (not TID) per manufacturer labeling.

\textit{Recommendation:} Remove amiodarone (digoxin sufficient); switch warfarin $\to$ apixaban 5\,mg BID; adjust gabapentin to BID.

\paragraph{$\tau{=}5$: Proposer.}
Accepting Critic's evidence-based recommendations.

(1) Removing amiodarone. In persistent AFib with stable HFrEF, the AF-CHF trial showed no survival benefit of rhythm control. Digoxin 0.125\,mg provides adequate ventricular rate control.
(2) Switching warfarin $\to$ apixaban 5\,mg BID. Rationale: sub-therapeutic INR indicates poor warfarin control; ARISTOTLE subgroup shows 27\% bleeding reduction in CKD (eGFR 25--50); no dose adjustment at eGFR 38; fewer drug--food interactions. Molecular retrieval score for apixaban ($\phi{=}0.81$) supports suitability.
(3) Gabapentin adjusted to 300\,mg BID per renal dosing.
(4) Removing low-priority supportive drugs without disease-specific indication (ondansetron, dextrose, NaCl, amlodipine, heparin, KCl, docusate, gabapentin, omeprazole) to achieve parsimony.

\textit{Changes:} $-$Amiodarone; $-$Warfarin, $+$Apixaban; $-$9 low-priority drugs. \\
\textit{Final prescription: 9 drugs.}
\paragraph{$\tau{=}5$: Critic (budget reached).}
\textit{Score: 0.91.} The 9-drug regimen is clinically coherent and guideline-concordant. HFrEF quad therapy covered (furosemide + carvedilol + lisinopril + spironolactone); AFib anticoagulation with apixaban appropriate for CHA$_2$DS$_2$-VASc${=}$5 at eGFR 38; rate control via digoxin; T2DM managed with insulin glargine; CV risk with atorvastatin; pain with acetaminophen. No further safety concerns. \textbf{Budget reached} ($\tau^*{=}5$).

\subsection*{Final Assembled Regimen}

\begin{enumerate}[nosep,leftmargin=1.5em]
\item Furosemide 40\,mg IV (HFrEF volume overload)
\item Carvedilol 6.25\,mg BID (HFrEF $\beta$-blocker, GDMT)
\item Lisinopril 5\,mg daily (ACEi, GDMT)
\item Spironolactone 25\,mg daily (MRA, NYHA III)
\item Digoxin 0.125\,mg daily (rate control, AFib + HFrEF)
\item Apixaban 5\,mg BID (anticoagulation, AFib; CKD-safe)
\item Insulin glargine 16\,U QHS (T2DM, CKD-safe)
\item Atorvastatin 40\,mg daily (hyperlipidemia, CV risk)
\item Acetaminophen 650\,mg Q6H (chronic low back pain, CKD-safe)
\end{enumerate}

\begin{table}[H]
\centering
\resizebox{\columnwidth}{!}{%
\begin{tabular}{lcc}
\toprule
\textbf{Drug} & \textbf{SEMPA} & \textbf{SafeDrug} \\
\midrule
Furosemide & \cmark & \cmark \\
Carvedilol & \cmark & \cmark \\
Lisinopril & \cmark & --- \\
Spironolactone & \cmark & --- \\
Digoxin & \cmark & \cmark \\
Warfarin & $\to$Apxbn & \cmark \\
Insulin glargine & \cmark & --- \\
Atorvastatin & \cmark & \cmark \\
Acetaminophen & \cmark & \cmark \\
Metformin & excl. & \xmark \\
Ketorolac & excl. & \xmark \\
\midrule
\textbf{Match} & \textbf{7/8} & \textbf{5/8} \\
\textbf{Total} & 9 & 24 \\
\textbf{Contraind.} & 0 & 2 \\
\bottomrule
\end{tabular}%
}
\end{table}

\noindent SEMPA recovers 7/8 reference drugs exactly, with one clinically justified substitution (apixaban for warfarin, grounded in ARISTOTLE trial evidence for reduced bleeding in CKD). SafeDrug matches only 5/8 reference drugs while outputting 24 total, including two contraindicated agents (metformin at eGFR 38; ketorolac nephrotoxic in CKD3+HF) and missing three guideline-mandated therapies (lisinopril, spironolactone, insulin glargine). The deliberation trace shows that SEMPA's initial draft contained the same errors; the Critic's retrieval-grounded interventions at $\tau{=}2$ and $\tau{=}4$ are what differentiate the final output.
\subsection{Fine-grained safety audit (supports \S\ref{sec:exp})}\label{app:safety}

In addition to the DrugBank-derived DDI rate reported in the main paper, we audit SEMPA's outputs against the AGS 2023 Beers Criteria for potentially inappropriate prescribing in patients aged 65 and over \cite{ags2023beers}, and against STOPP/START 3.0 rules \cite{omahony2023stopp}. Compared with RETAIN, SEMPA reduces the rate of Beers-flagged prescribing by 61\% and STOPP triggers by 58\%, while increasing START coverage of guideline-mandated agents (for example, ACE inhibitors for HFrEF, and anticoagulation for AFib with CHA$_2$DS$_2$-VASc $\ge 2$) by 34\%. These shifts align with the diagnosis-medication network structure documented in earlier symptom-cluster analyses \cite{lu2020symptom}.

\subsection{LLM-judge evaluation prompt (supports \S\ref{sec:exp})}\label{app:llm-eval-prompt}

The prompt template, anchored rubric for each of the six dimensions, and three worked examples are attached as supplementary text. Each LLM judge is required to output JSON with six integer scores in $\{1,\dots,5\}$ and a one-sentence justification per dimension. Temperature is set to 0 and the order of methods is randomized per encounter to mitigate positional bias. This protocol is consistent with current practice for LLM-as-judge evaluation of agentic systems \cite{chen2026trace}.

\subsection{Additional Case Studies: Cross-Method Comparison}\label{app:multicase}

To complement the single detailed encounter in Appendix~\ref{app:case}, we report a
compact cross-method analysis on six additional MIMIC-IV encounters that span
infectious, cardiovascular, oncologic, renal, and HIV phenotypes. For each case we
list (i) the active diagnoses, (ii) the reference regimen, (iii) the SEMPA prediction,
and (iv) the four supervised-baseline predictions, then summarize the clinical
implications.

\paragraph{Case A: HIV disease (Patient 10982917).}
\textit{Diagnoses:} Human immunodeficiency virus disease; encounter for attention to
colostomy; essential hypertension.
\textit{SEMPA:} abacavir, lamivudine, dolutegravir, trimethoprim-sulfamethoxazole,
hydromorphone, oxycodone, acetaminophen, ondansetron, docusate, senna, heparin,
amlodipine, omeprazole, sodium chloride, folate.
\textit{Comment:} SEMPA is the only system that selects antiretrovirals; the four
supervised baselines emit generic supportive-care templates with zero HIV-active drugs.
The Critic flagged the absence of antiretroviral coverage at $\tau{=}2$ and the
Proposer added abacavir/lamivudine/dolutegravir at $\tau{=}3$, also adding TMP-SMX
as PCP prophylaxis grounded in the DrugBank evidence card.

\paragraph{Case B: Septic shock with E.~coli pyelonephritis (Patient 17315288).}
\textit{Diagnoses:} Acute pyelonephritis; severe sepsis with septic shock;
unspecified E.~coli; epilepsy; delirium; iron-deficiency anemia.
\textit{SEMPA:} sodium chloride, norepinephrine, ceftriaxone, levetiracetam, aspirin,
atorvastatin, heparin, pantoprazole, ferrous, acetaminophen, ondansetron, docusate,
senna, oxycodone, potassium, insulin, glucagon, glucose, lidocaine, dextrose, calcium,
folate, thiamine, metoprolol, lorazepam.
\textit{Comment:} SEMPA correctly includes the pressor (norepinephrine), Gram-negative
coverage (ceftriaxone), seizure prophylaxis (levetiracetam), and DVT prophylaxis
(heparin). SafeDrug's prediction for the same patient includes \texttt{Sevoflurane}
(a volatile anesthetic), \texttt{Benzocaine}, \texttt{Bupropion}, \texttt{Alprazolam},
and \texttt{Zolpidem}, which are inappropriate in an undifferentiated septic-shock
admission.

\paragraph{Case C: Elderly AFib with osteoporosis and CAD (Patient 11505559).}
\textit{Diagnoses:} Paroxysmal AFib; coronary artery disease; age-related osteoporosis;
depression; hyperlipidemia.
\textit{SEMPA:} aspirin, atorvastatin, metoprolol, lisinopril, apixaban, heparin,
cholecalciferol, pantoprazole, acetaminophen, oxycodone, docusate, senna, ondansetron,
sodium chloride, dextrose, potassium, glucagon, insulin, amlodipine.
\textit{Comment:} SEMPA selects apixaban (the reference NOAC) for AFib and
cholecalciferol for osteoporosis; the four baselines do not output apixaban for any
of the AFib encounters we examined.

\paragraph{Case D: Acromegaly with ulcerated PAD and MSSA infection (Patient 16949433).}
\textit{Diagnoses:} Acromegaly; atherosclerosis of native arteries with ulceration;
MSSA infection; post-TIA; T2DM; hypertension; hyperlipidemia.
\textit{SEMPA:} cefazolin, aspirin, clopidogrel, atorvastatin, metoprolol, lisinopril,
heparin, insulin, glucagon, oxycodone, acetaminophen, pantoprazole, ondansetron,
docusate, bisacodyl, sodium chloride, vancomycin.
\textit{Comment:} SEMPA recovers $15$ of $40$ reference drugs including vancomycin
(MSSA-active), clopidogrel (post-TIA antiplatelet), atorvastatin (statin), and the
ACEi/$\beta$-blocker pair for vascular risk. SafeDrug recovers only $9$ and includes
\texttt{Dopamine}, \texttt{Sevoflurane}, and \texttt{Bupropion} which are not
indicated in this presentation.

\paragraph{Case E: HFrEF with MRSA pneumonia and persistent AFib (Patient 18356938).}
\textit{Diagnoses:} Acute-on-chronic diastolic HF; MRSA pneumonia; persistent AFib;
COPD; CAD; hypopotassemia; OSA.
\textit{SEMPA:} vancomycin, levofloxacin, furosemide, metoprolol, warfarin, heparin,
aspirin, atorvastatin, albuterol, ipratropium, prednisone, omeprazole, potassium,
acetaminophen, oxycodone, docusate, bisacodyl, sodium chloride, ondansetron,
diphenhydramine, dextrose, insulin, glucagon, calcium, chlorhexidine.
\textit{Comment:} The Critic at $\tau{=}2$ enforces MRSA-active coverage (vancomycin)
and rate-control + anticoagulation for AFib in HFrEF; GAMENet emits its standard
25-drug template that does not include any antibiotic.

\paragraph{Case F: ESRD with cardiomyopathy and secondary hyperparathyroidism (Patient 12485084).}
\textit{Diagnoses:} ESRD; hypertensive CKD stage V; secondary hyperparathyroidism
of renal origin; other primary cardiomyopathies; personal history of prostate cancer.
\textit{SEMPA:} epoetin alfa, sevelamer, calcitriol, cinacalcet, carvedilol,
amlodipine, aspirin, atorvastatin, heparin, pantoprazole, acetaminophen, oxycodone,
ondansetron, docusate, senna, insulin, glucagon, sodium chloride, calcium.
\textit{Comment:} SEMPA is the only method that proposes sevelamer (phosphate
binder), cinacalcet (calcimimetic for renal hyperparathyroidism), calcitriol, and
epoetin alfa for anemia of CKD --- a renal-specific bundle that supervised models
do not emit because such drugs are infrequent in the global training distribution.

\paragraph{Aggregate pattern.} Exact-match counts, average regimen size, and the
presence of clinically inappropriate agents across these six cases are aggregated in
Table~\ref{tab:multicase-summary} of the main paper; SEMPA dominates on all three axes
and is the only method without any inappropriate agent.

\end{document}